\documentclass[11pt]{article}

\usepackage[preprint]{acl}

\usepackage{times}
\usepackage{latexsym}

\usepackage[T1]{fontenc}

\usepackage[utf8]{inputenc}

\usepackage{microtype}

\usepackage{inconsolata}
\usepackage{hyperref}
\usepackage{url}
\usepackage{booktabs}
\usepackage{hyperref}
\usepackage{url}
\usepackage{todonotes}

\usepackage{tabularx}
\usepackage{booktabs}
\usepackage{wrapfig}
\usepackage{booktabs}
\usepackage{threeparttable}
\usepackage{graphicx}
\usepackage{subcaption}
\usepackage{float}
\usepackage{multirow}
\usepackage[T1]{fontenc}
\usepackage{listings}
\usepackage{amsmath} 

\usepackage{lineno}

\definecolor{darkblue}{rgb}{0, 0, 0.5}
\hypersetup{colorlinks=true, citecolor=darkblue, linkcolor=darkblue, urlcolor=darkblue}

\title{Markovian Generation Chains in Large Language Models}

\author{
Mingmeng Geng\thanks{Equal contribution.} \\
ENS-PSL \& CNRS-Lattice \\
\texttt{mingmeng.geng@ens.psl.eu} \\
\And
\qquad Amr Mohamed\footnotemark[1] \\
\qquad MBZUAI \& Ecole Polytechnique \\
\qquad \texttt{amr.mohamed@mbzuai.ac.ae}
\And
Guokan Shang \\
MBZUAI
\AND
Michalis Vazirgiannis \\
MBZUAI \&
Ecole Polytechnique
\And
Thierry Poibeau \\
ENS-PSL \&
CNRS-Lattice
}

\begin{document}

\maketitle
\begin{abstract}

The widespread use of large language models (LLMs) raises an important question: how do texts evolve when they are repeatedly processed by LLMs? In this paper, we define this iterative inference process as \textit{Markovian generation chains}, where each step takes a specific prompt template and the previous output as input, without including any prior data. In iterative rephrasing and round-trip translation experiments, the output either converges to a small recurrent set or continues to produce novel sentences within a finite horizon. Through sentence-level Markov chain modeling and analysis of simulated data, we show that iterative process can either increase or reduce the diversity depending on factors such as the temperature parameter and the initial input sentence. These results offer valuable insights into the dynamics of iterative LLM inference and their implications for multi-agent LLM systems.\footnote{Data and code are available at: \href{https://github.com/PolyGN/Markovian-Generation-Chains-in-Large-Language-Models}{https://github.com/PolyGN/Markovian-Generation-Chains-in-Large-Language-Models}}
\end{abstract}

\section{Introduction}

Large language models (LLMs) are used in a wide range of downstream tasks, such as translation and rewriting. As LLM-generated content becomes more prevalent, the likelihood that such content will be iteratively reprocessed increases. This motivates a practical but underexplored question: \emph{how do texts evolve under repeated LLM reprocessing?}

We refer to this process as \textit{Markovian generation chains in LLMs}: the input consists only of a specific prompt template and the output of the previous inference, without including any prior memory. Although such recursive reuse arises naturally in iterative translation and repeated rephrasing workflows~\citep{perez2025llms,mohamed2025llm}, it currently lacks a standard formalization and well-defined metrics for characterizing these outcomes. While LLM outputs are largely determined by the input, the model~\citep{jang2016categorical} and the hardware~\citep{he2025nondeterminism} may also introduce variation. To this end, we propose modeling this process using Markov chains.

We treat each \emph{sentence} as the primary unit of analysis and focus on the evolution in the iterative procedure, e.g., quantifying diversity based on how many \emph{distinct sentences} emerge over the repeated process. \citet{zekri2024large} analyze a single inference at the token level using Markov chains. Markov chains have also been used in other studies on LLMs, such as the reasoning process~\citep{teng2025atom}. A sentence is a combination of tokens and may carry more information and different patterns, and our focus is on the relationship between inference iterations. Moreover, we apply additional metrics to measure changes in word usage and semantic drift. 

\begin{figure*}[t]
    \centering
    \includegraphics[width=0.98\textwidth]{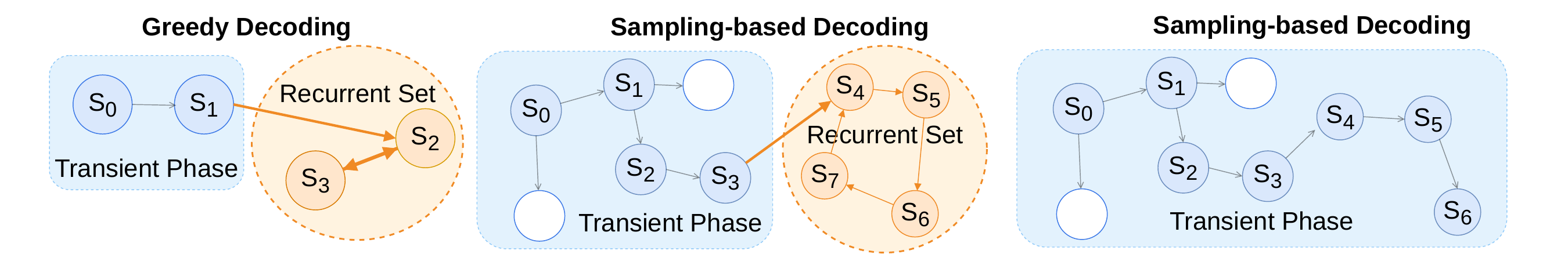}
    \vspace{-0.5em}
    \caption{
    Illustration of iterative LLM reprocessing (\textit{Markovian generation chains}).
    Each node denotes the output after iteration $i$.
    Under \textbf{greedy decoding}, chains typically enter fixed points or short cycles, limiting sentence-level diversity.
    Under \textbf{sampling-based decoding}, stochasticity may yield longer transients and more distinct outputs.}
    \label{fig:experiments_illustration}
    \vspace{-1em}
\end{figure*}

The accumulation of random perturbations and systematic biases can lead to different regimes: equilibrium, periodic behavior, divergence, or random walk. The prevalence of these behaviors depends on the model, the decoding configuration, and the seed input sentence; in particular, sampling-based decoding typically increases exploration and prolongs pre-recurrence phases.  Figure~\ref{fig:experiments_illustration} provides an illustration of these behaviors under greedy versus sampling-based decoding.

These iterative dynamics matter beyond controlled settings because LLM-mediated text can propagate through downstream communication and decision pipelines. Prior work suggests that LLMs may reshape collective intelligence~\citep{burton2024large}, and that repeated generation can introduce information distortion even when prompts request meaning preservation~\citep{mohamed2025llm}. 

The main contributions of this paper are as follows:
\begin{itemize}
    \item We introduce the concept of a Markovian generative chain. In our experiments, text produced by an LLM via paraphrasing or translation is recursively reused as input, approximating multi-step and multi-user reprocessing workflows in the real world. 
    \item We propose constructing a Markov chain at the sentence level for observation and analysis. The sentence-level framing supports a natural mapping between an input sentence and the set of plausible output sentences, and it could be extended to paragraph-level simulations as an ablation.
    \item Simulation results and theoretical analysis show that text repeatedly processed by LLMs can either diverge or fall into cycles. Its diversity may increase or decrease, depending on the input and the model.
\end{itemize}

\section{Methodology}
\label{sec:methodology}

This section formalizes iterative processes as \textit{Markovian generation chains} and introduces the tools used to analyze the empirical regimes in our experiments.

\subsection{Iterative reprocessing as a Markovian generation chain}
\label{subsec:method_mgc}

Let $s^{(0)}$ denote an initial \emph{sentence}. Fix (i) a model $M$, (ii) a prompt template $\rho$, and (iii) a decoding configuration $d$ (e.g., greedy decoding, or sampling-based decoding with specified temperature/top-$p$). One step of iterative reprocessing defines a stochastic transformation operator $\mathcal{T}_{M,\rho,d}$ that maps the text-unit surface string to a distribution over surface strings:
\begin{equation}
s^{(t+1)} \sim \mathcal{T}_{M,\rho,d}(\,\cdot \mid s^{(t)}),
\enspace t=0,\dots,T-1.
\label{eq:mgc_step}
\end{equation}

Equivalently, $\mathcal{T}_{M,\rho,d}$ defines a time-homogeneous Markov kernel $P_{M,\rho,d}(s' \mid s)$ over sentences. We restrict the state space by requiring each iteration to output exactly one sentence. Further details are explained in Appendix~\ref{appendix_method_decoding}. 

\paragraph{Operational Markovian assumption.}
Iterations are independent model calls conditioned only on $\rho$ and $s^{(t)}$; no history or latent state is carried between steps. Thus $\{s^{(t)}\}_{t=0}^{T}$ is Markovian, with any system nondeterminism absorbed into $P_{M,\rho,d}$.

\paragraph{Structured operators via composition.}
A single iteration can be written as a composition of prompted calls. For round-trip translation with bridge language $\ell$, one iteration composes two translations (EN$\to \ell$ then $\ell\to$EN). Let $\mathcal{T}^{\text{EN}\to \ell}_{M,\rho,d}$ and $\mathcal{T}^{\ell\to \text{EN}}_{M,\rho,d}$ denote the induced operators for each direction under the same fixed setup. The resulting \textsc{EN}-to-\textsc{EN} two-step operator is
\begin{equation}
\mathcal{T}^{\text{EN}\to \ell \to \text{EN}}_{M,\rho,d}
\;:=\;
\mathcal{T}^{\ell\to \text{EN}}_{M,\rho,d}
\circ
\mathcal{T}^{\text{EN}\to \ell}_{M,\rho,d},
\label{eq:rt_operator}
\end{equation}
which induces a Markov chain on English sentences. At the kernel level, this composition corresponds to the standard product of Markov kernels:

\begin{equation}
\begin{split}
&P^{\text{EN}\to \ell \to \text{EN}}_{M,\rho,d}(s' \mid s) \\
&\quad = \sum_{u \in \mathcal{S}_{\ell}}
      P^{\ell\to \text{EN}}_{M,\rho,d}(s' \mid u)\,
      P^{\text{EN}\to \ell}_{M,\rho,d}(u \mid s),
\end{split}
\label{eq:rt_kernel}
\end{equation}
where $\mathcal{S}_{\ell}$ denotes the (conceptual) state space of sentences in language $\ell$.

\subsection{Sentence-level Markov chain formulation}
\label{subsec:method_sentence_mc}

We model iterative reprocessing at the \emph{sentence} level, treating sentence strings as discrete states and the prompted model as a transition operator. For conceptual clarity, we consider finite (but extremely large) state spaces.

\paragraph{State spaces.}
For translation between languages $l_A$ and $l_B$, let $\mathcal{S}_A = \{a_1,\ldots,a_m\},\ \mathcal{S}_B = \{b_1,\ldots,b_n\}$. For rephrasing, the chain remains within a single language, i.e., $\mathcal{S}_A \to \mathcal{S}_A$.

\paragraph{Transition matrices.}
Fix $(M,\rho,d)$. For translation from $l_A$ to $l_B$, define the sentence-level transition matrix

\begin{equation}
\begin{split}
[\mathbf{P}^{A\to B}_{M,\rho,d}]_{ij}
&:= \Pr\!\left(b_j \mid a_i;\, M,\rho,d\right) \\
&= P^{A\to B}_{M,\rho,d}(b_j \mid a_i).
\end{split}
\end{equation}
For round-trip translation $l_A \to l_B \to l_A$, the induced within-$\mathcal{S}_A$ transition matrix is the product
\begin{equation}
\mathbf{P}^{A\to B\to A}_{M,\rho,d}
=
\mathbf{P}^{A\to B}_{M,\rho,d}\,
\mathbf{P}^{B\to A}_{M,\rho,d},
\label{eq:rt_matrix}
\end{equation}
mirroring the kernel composition in Eq.~(\ref{eq:rt_kernel}). If $X^{(0)}$ denotes an initial distribution over $\mathcal{S}_A$ represented as a row vector, then after $n$ iterations
\begin{equation}
X^{(n)} = X^{(0)} \left(\mathbf{P}^{A\to B\to A}_{M,\rho,d}\right)^{n}.
\end{equation}
For iterative rephrasing, we use the analogous within-language matrix $\mathbf{P}_{M,\rho,d}$ on $\mathcal{S}_A$.

\subsection{Regimes under iteration: recurrent classes and transients}
\label{subsec:method_regimes}

Empirically, iterative reprocessing exhibits two finite-horizon behaviors: (i) \emph{early exact recurrence} (a fixed point or short cycle) and (ii) \emph{long pre-recurrence phases} (continued production of novel surface forms within the iteration budget). We characterize these regimes using finite-horizon recurrence statistics computed on exact string equality.

\paragraph{Asymptotic structure (conceptual).}
For a finite state space, $\mathbf{P}$ can be permuted into the standard block form with recurrent classes $\{C_i\}$ and a transient block $\mathbf{P}_{\mathrm{tr}}$ (Eq.~\ref{eq:blockP}); early recurrence corresponds to fast entry into a small recurrent class, while long pre-recurrence corresponds to trajectories that remain in transient regions over the finite horizon.
\begin{equation}
\mathbf{P}=
\begin{bmatrix}
\mathbf{P}_1 & \mathbf{0} & \cdots & \mathbf{0} & \mathbf{0}\\
\mathbf{0} & \mathbf{P}_2 & \cdots & \mathbf{0} & \mathbf{0}\\
\vdots & \vdots & \ddots & \vdots & \vdots\\
\mathbf{0} & \mathbf{0} & \cdots & \mathbf{P}_c & \mathbf{0}\\
\mathbf{Q}_1 & \mathbf{Q}_2 & \cdots & \mathbf{Q}_c & \mathbf{P}_{\mathrm{tr}}
\end{bmatrix}\!,
\label{eq:blockP}
\end{equation}
where $\mathbf{P}_i$ governs motion within $C_i$, $\mathbf{Q}_i$ captures transitions from transient states into $C_i$, and $\mathbf{P}_{\mathrm{tr}}$ governs transient-to-transient motion. Since the empirical trajectories are finite, the chain may not exhibit observable recurrence within the chosen horizon. 

\paragraph{Finite-horizon recurrence statistics.}
We treat ``exact'' equality as literal string equality on model outputs. Given a trajectory $\{s^{(t)}\}_{t=0}^{T}$, the first recurrence time is

\begin{equation}
\begin{split}
\tau_T = \min\bigl\{ t \in \{1,\dots,T\} : \qquad & \\
\exists j<t \text{ s.t. } s^{(t)} = s^{(j)} \bigr\}&,
\end{split}
\label{first_recurrence_time}
\end{equation}
with $\tau_T=T+1$ if no exact repeat occurs within the horizon. Small $\tau_T$ corresponds to early entry into a fixed point or short cycle; large $\tau_T$ (or $T+1$) corresponds to a long pre-recurrence phase within $T$ steps.

\paragraph{Effect of decoding on pre-recurrence behavior.}
Decoding parameters in $d$ (e.g., temperature/top-$p$) affect $\mathbf{P}$ by redistributing transition mass. Sampling-based decoding typically increases probability assigned to lower-ranked continuations, enlarging the set of accessible next-step realizations and thereby increasing $\tau_T$ on average, i.e., prolonging pre-recurrence phases relative to greedy decoding.

\subsection{Information-theoretic tools for iterative reprocessing}
\label{subsec:method_info}

We record two standard properties of the induced kernel $P$ — entropy behavior and Kullback--Leibler (KL) contraction — that bear on diversity and on convergence under iteration.

\paragraph{Entropy under doubly stochastic kernels.}
Let $X$ be a distribution over sentences and $\mathbf{P}$ a row-stochastic transition matrix (the kernel induced by $(M,\rho,d)$). The sentence entropy is
\begin{equation}
\mathrm{H}(X)=-\sum_s X(s)\log X(s)\,. \label{eq:hx}
\end{equation}

If $\mathbf{P}$ is doubly stochastic, then $\mathrm{H}(X\mathbf{P})\ge \mathrm{H}(X)$; otherwise (as for prompted LLM kernels) the entropy may increase or decrease depending on $X$ and $\mathbf{P}$.

\paragraph{Data processing and KL contraction.}
For distributions $X,Y$ over sentences, KL divergence contracts under a
stochastic matrix:
\begin{equation}
D_{\mathrm{KL}}(X\mathbf{P}\,\|\,Y\mathbf{P}) \le D_{\mathrm{KL}}(X\,\|\,Y)\,,
\end{equation}
Proof is provided in Appendix~\ref{appendix_KL}. If $\pi$ is stationary ($\pi\mathbf{P}=\pi$), then $D_{\mathrm{KL}}(X\mathbf{P}^n\|\pi)$ is non-increasing in $n$. But the convergence also depends on the existence of a single recurrent class and aperiodicity, neither of which is guaranteed in this setting.

\subsection{Measurement and evaluations}
We used the following methods for measurement and evaluation:
\begin{itemize}
    \item \emph{Diversity:} sentence diversity (the number of distinct sentences) and token diversity (the ratio of unique tokens to total tokens).
    \item \emph{Drift:} METEOR~\citep{banerjee2005meteor}, ROUGE-1~\citep{lin2004rouge}, BLEU~\citep{papineni2002bleu}, TF--IDF cosine (2--4 grams), ChrF~\citep{popovic2015chrf}, BERTScore F1~\citep{zhang2019bertscore}, anchor preservation (F1 overlap of factual/lexical anchors), and compression ratio (token-count ratio between adjacent iterations), computed stepwise ($s^{(t)}$ vs.\ $s^{(t-1)}$) and cumulative ($s^{(t)}$ vs.\ $s^{(0)}$).
    \item \emph{Input sensitivity:} correlation of seed length with $U$ (Pearson $r$ and linear regression), where $U$ counts distinct surface strings (including minor variants).
\end{itemize}

\section{Experimental Setup}
\label{sec:experimental_setup}

\paragraph{Data}
We use three corpora spanning distinct domains: \textit{BookSum}~\citep{kryscinski2022booksum}, \textit{ScriptBase-alpha}~\citep{gorinski2015movie}, and \textit{(BBC) News2024}~\citep{li2024latesteval}. From each dataset, we randomly sample 150 documents and extract the first sentence to serve as the seed $s^{(0)}$.

\paragraph{Models and baselines.}
We evaluate instruction-tuned open-weight models: \textit{Mistral-7B-Instruct}~\citep{jiang2023mistral}, \textit{Llama-3.1-8B-Instruct}~\citep{dubey2024llama}, and \textit{Qwen2.5-7B-Instruct}~\citep{qwen2}. We also include \textit{GPT-4o-mini}~\citep{hurst2024gpt} via its API. For round-trip translation, we compare prompted LLM translation to \textit{Google Translate (v3)} (API), a near-deterministic production MT baseline for fixed inputs. The latest DeepSeek V4 is also used in the ablation experiments.

\paragraph{Decoding configurations and prompts.}
We compare two decoding regimes. Under \textbf{greedy decoding}, tokens are selected by argmax at each step, close to a deterministic mapping. Under \textbf{sampling-based decoding}, we sample with fixed hyperparameters for comparability (unless otherwise stated, $\tau=0.7$, top-$p=0.9$). When the stack exposes an explicit random seed, we fix it per chain; otherwise (e.g., APIs) we treat each chain as a stochastic realization and aggregate over many runs. We use $\rho_{\mathrm{reph}}$ for rephrasing and $\rho_{\mathrm{tr}}$ for each translation direction. Full templates (including alternation variants) are in Appendix~\ref{appendix_prompts}.

\section{Results}
\subsection{Main results and findings}

\begin{table*}[t]
\centering
\small
\begin{tabularx}{\textwidth}{>{\raggedright\arraybackslash}p{1cm} X X}
\toprule
Round &  Qwen2.5-7B-Instruct & Llama-3.1-8B-Instruct \\
\midrule
0 & We \textit{begin} with a prologue. & We \textit{begin} with a prologue. \\
1 & We \textbf{start} with a prologue. & \underline{The story} \textbf{commences} with a prologue. \\
2 & We \textit{begin} with a prologue. & \underline{The narrative} \textit{begins} with a prologue. \\
3 & We \textbf{start} with a prologue. & \underline{The story} \textbf{commences} with a prologue. \\ 
4 & We \textit{begin} with a prologue. & \underline{The narrative} \textit{begins} with a prologue. \\
5 & We \textbf{start} with a prologue. & \underline{The story} \textbf{commences} with a prologue. \\
6 & We \textit{begin} with a prologue. & \underline{The narrative} \textit{begins} with a prologue.\\
\bottomrule
\end{tabularx}
\vspace{-0.5em}
\caption{Illustrative greedy-decoding trajectories under iterative rephrasing. 
Qwen2.5-7B enters a 2-cycle after one step, while Llama-3.1-8B alternates between two near-paraphrases with different lexical realizations.}
\label{rephrasing_example}
\vspace{-0.5em}
\end{table*}
\begin{figure*}[t]
    \centering
    \includegraphics[width=\textwidth]{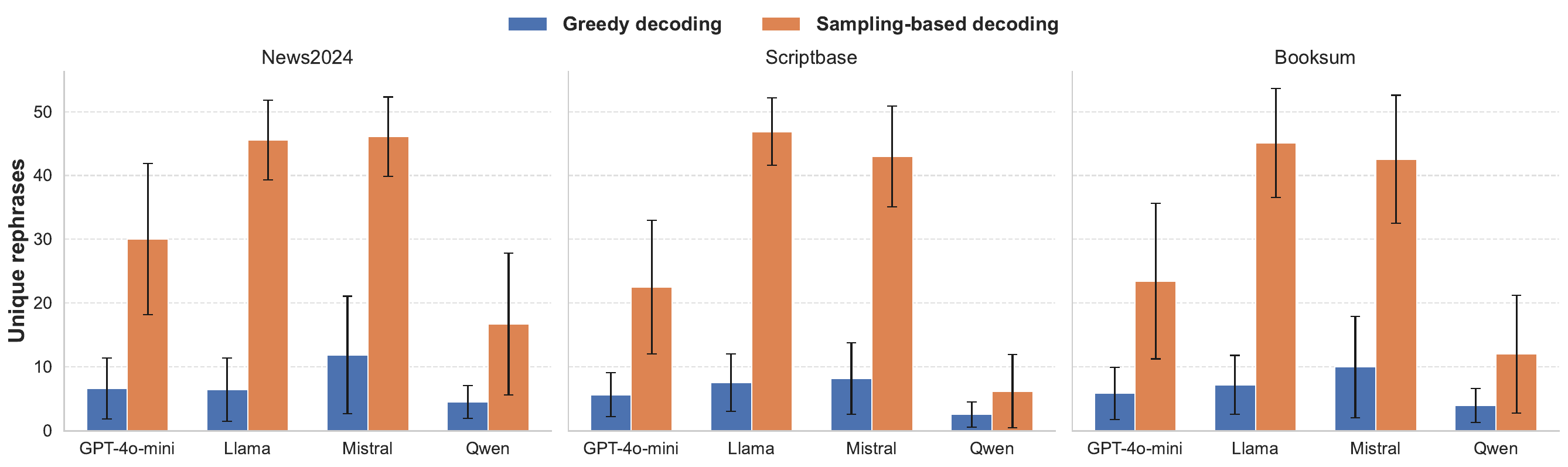}
    \vspace{-0.5em}
    \caption{Average number of unique paraphrases generated in 50 iterations across three datasets, comparing four instruction-tuned LLMs: GPT-4o-mini, Llama-3.1-8B, Mistral-7B, and Qwen-2.5-7B. Results are shown for greedy decoding (blue) and sampling-based decoding (orange). Error bars represent one standard deviation.} 
    \label{fig:barplot_rephrasings}
\end{figure*}

\begin{figure*}[t]
    \centering
    \includegraphics[width=\textwidth]{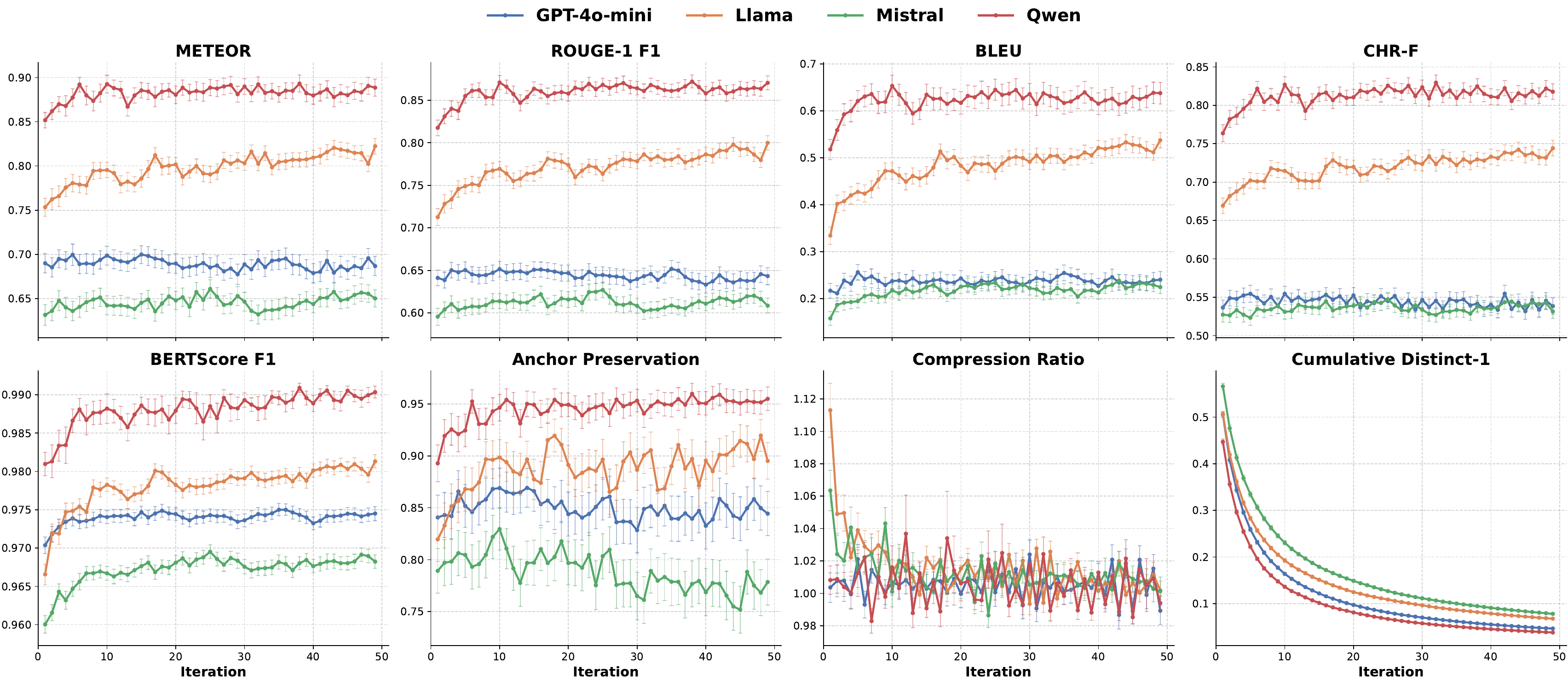}
    \caption{Evolution of text similarity metrics across 50 rephrasing iterations for the BookSum dataset using sampling-based decoding. Each iteration compares the current rephrased text against the previous iteration's text as reference.}
    \vspace{-1em}

    \label{fig:booksum_models_comparison_subplot_sampling}
\end{figure*}

Table~\ref{rephrasing_example} provides an illustrative greedy-decoding trajectory and Figure~\ref{fig:barplot_rephrasings} reports sentence-level diversity over $T{=}50$ iterations across datasets and models. Under greedy decoding, trajectories typically enter small recurrent sets (fixed points or short cycles) after few steps, yielding repeated surface strings or short alternations among near-paraphrases. The observed cycle structure and lexical realizations vary by model, indicating model-specific attractors under a fixed prompting setup. For sampling-based decoding, trajectories exhibit longer pre-recurrence phases. Across all datasets, greedy decoding produces substantially fewer unique paraphrases, consistent with rapid entry into recurrent behavior. Sampling yields markedly larger support sizes with substantial heterogeneity across models and domains, suggesting that both the induced kernel (model/decoding) and the seed distribution modulate time-to-recurrence within the finite horizon. In addition, different LLMs also perform differently, suggesting that they may have different attractors. 

To characterize local dynamics along trajectories, we employ eight different metrics (e.g., METEOR and BERTScore) to measure and compare lexical and semantic drift between successive iterations ($s^{(t)}$ vs.\ $s^{(t-1)}$). With sampling-based decoding, the evolution of sentences across iterations is non-monotonic at either the textual or semantic level, as illustrated in Figure~\ref{fig:booksum_models_comparison_subplot_sampling} and Appendix Figures~\ref{fig:news2024_models_comparison_subplot_sampling} and \ref{fig:scriptbase_models_comparison_subplot_sampling}. Under greedy decoding, these stepwise similarity scores rapidly plateau (Appendix Figures~\ref{fig:booksum_models_comparison_subplot_greedy}, \ref{fig:news2024_models_comparison_subplot_greedy} and~\ref{fig:scriptbase_models_comparison_subplot_greedy}), consistent with confinement to a small recurrent set rather than continued exploration. Differences in plateau levels across models reflect variation in the lexical and semantic distance between recurrent paraphrase variants.

\subsection{Parameters and inputs}
The behavior of iterative reprocessing depends on both the \emph{decoding configuration} and the \emph{initial input}. Increasing the temperature generally flattens the distribution and increases the probability of selecting lower-ranked tokens, thereby expanding the set of plausible continuations. Supplementary results in Figure~\ref{fig:booksum_models_comparison_subplot_sampling} and Appendix  Figures~\ref{fig:news2024_models_comparison_subplot_sampling}--\ref{fig:scriptbase_models_comparison_subplot_greedy} indicate that sampling-based decoding settings tend to delay or prevent rapid entry into small recurrent sets, yielding longer transients with fewer exact repetitions.

The initial input affects the size of the accessible paraphrastic neighborhood and thus the diversity observed along an iterative chain. Table~\ref{tab:length_diversity_r} reports the association between seed length (words) and sentence-level diversity (the number of distinct outputs over $T=50$ iterations) across models, decoding regimes, and datasets. Correlations are generally positive but heterogeneous: sampling often amplifies the effect (notably on \textsc{BookSum} and \textsc{News2024}), while \textsc{ScriptBase} is weaker or inconsistent. Hence, the type of text may also affect the experimental results. Full correlation and regression results are given in Appendix~\ref{app:appendix_addition_results} (Table~\ref{tab_correlation}).
\begin{table}[t]
\small
\begin{tabular*}{\columnwidth}{@{\extracolsep{\fill}} lccc @{}}
\toprule
\textbf{Model (Decoding)} &
\multicolumn{1}{c}{\textbf{BookSum}} &
\multicolumn{1}{c}{\textbf{ScriptBase}} &
\multicolumn{1}{c}{\textbf{News2024}} \\
\midrule
Llama (greedy)   & 0.17 & -0.17 & 0.19 \\
Llama (sampling)   & 0.17 & 0.03  & 0.11 \\
Mistral (greedy) & 0.32 & 0.07  & 0.11 \\
Mistral (sampling) & 0.44 & -0.02 & 0.20 \\
Qwen (greedy)    & 0.35 & 0.09  & 0.20 \\
Qwen (sampling)    & 0.49 & 0.16  & 0.37 \\
GPT (greedy)     & 0.48 & 0.10  & 0.37 \\
GPT (sampling)     & 0.64 & 0.23  & 0.39 \\
\bottomrule
\end{tabular*}
\caption{Pearson correlation $r$ between seed length and the number of distinct outputs over $50$ iterations.} 
\label{tab:length_diversity_r}
\vspace{-1.5em}
\end{table}

\subsection{Ablations}

We perform a set of ablation studies to evaluate the robustness of the observed iterative regimes and to identify which elements of the chain specification most strongly affect recurrence and output diversity. Specifically, we vary: (i) the prompt template, (ii) prompt heterogeneity across iterations, (iii) the granularity of the input unit (sentences vs.\ paragraphs), and (iv) the task instantiation via round-trip translation.

\subsubsection{Sensitivity to prompts}

\begin{figure*}[t]
    \centering
    \includegraphics[width=\textwidth]{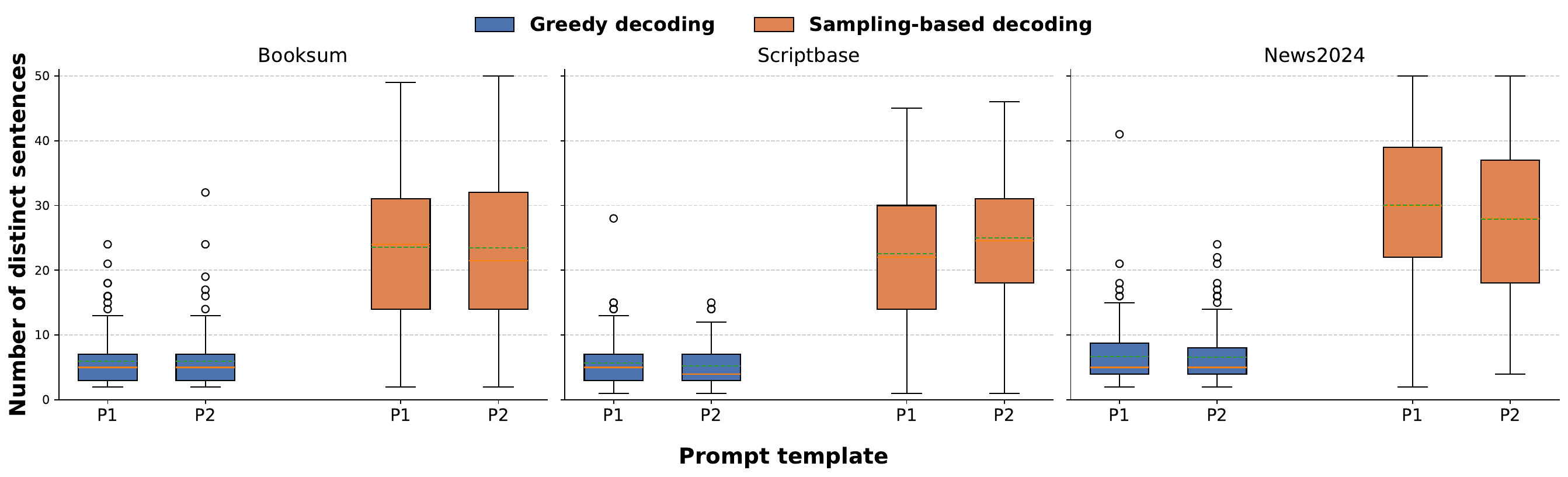}
    \vspace{-1.5em}
    \caption{Number of distinct sentences produced over 50 iterative rephrasings with \textit{GPT-4o-mini} under different settings. P1 and P2 correspond to Listings~\ref{prompt_example} and~\ref{prompt_ablation} in the Appendix. Boxes denote the interquartile range, and the center line indicates the median.} 
    \label{gpt4o_mini_output_version_prompt}
    \vspace{-1em}
\end{figure*}

We first assess the sensitivity of iterative rephrasing dynamics to the prompt template. Using \textit{GPT-4o-mini} as a representative model, we compare two prompt variants (P1 and P2; Listings~\ref{prompt_example} and~\ref{prompt_ablation}). As shown in Figure~\ref{gpt4o_mini_output_version_prompt}, the decoding regime is the primary driver within this prompt range: sampling-based decoding consistently yields a substantially larger number of distinct sentence realizations than greedy decoding, whereas the prompt variation induces comparatively smaller changes. For the prompts considered here, the induced operator is not altered sufficiently to dominate recurrence behavior largely determined by decoding.

\subsubsection{Alternating between different prompts and different LLMs}

\begin{figure*}[t]
    \centering
    \includegraphics[width=\textwidth]{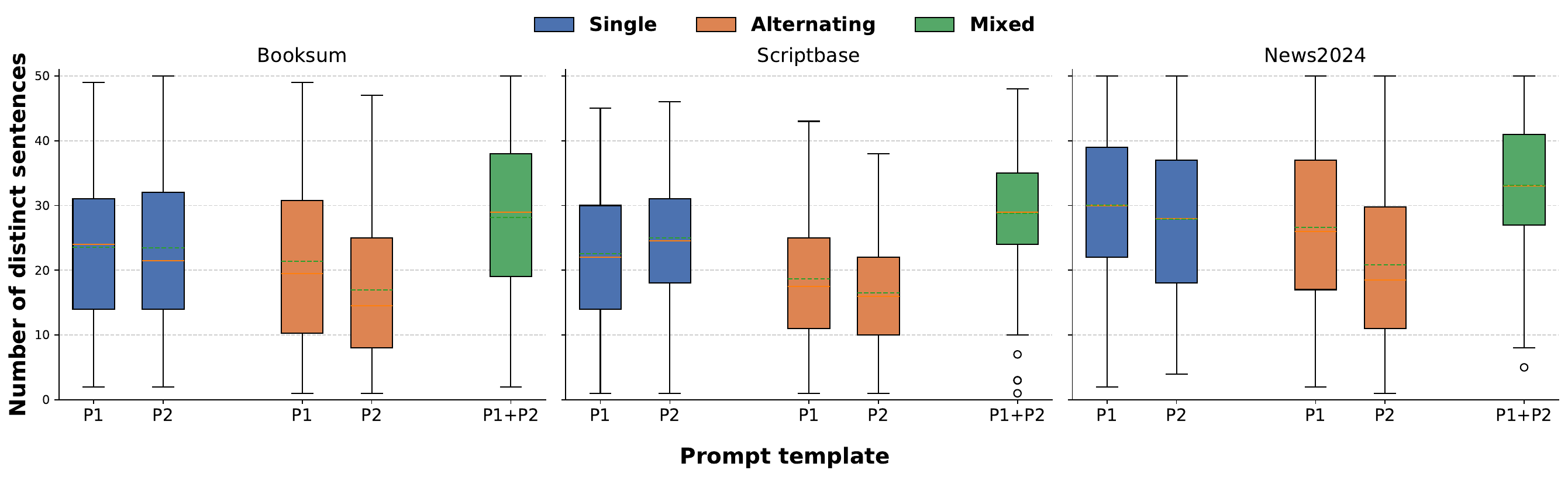}
    \vspace{-1.5em}
    \caption{Prompt heterogeneity across iterations (\textit{GPT-4o-mini}, sampling-based decoding). We compare a single fixed prompt against alternating prompts, and a mixed variant constructed from prompt halves.} 
    \label{prompt_mix}
    \vspace{-0.5em}
\end{figure*}

\begin{figure}[t]
    \centering
    \includegraphics[width=\columnwidth]{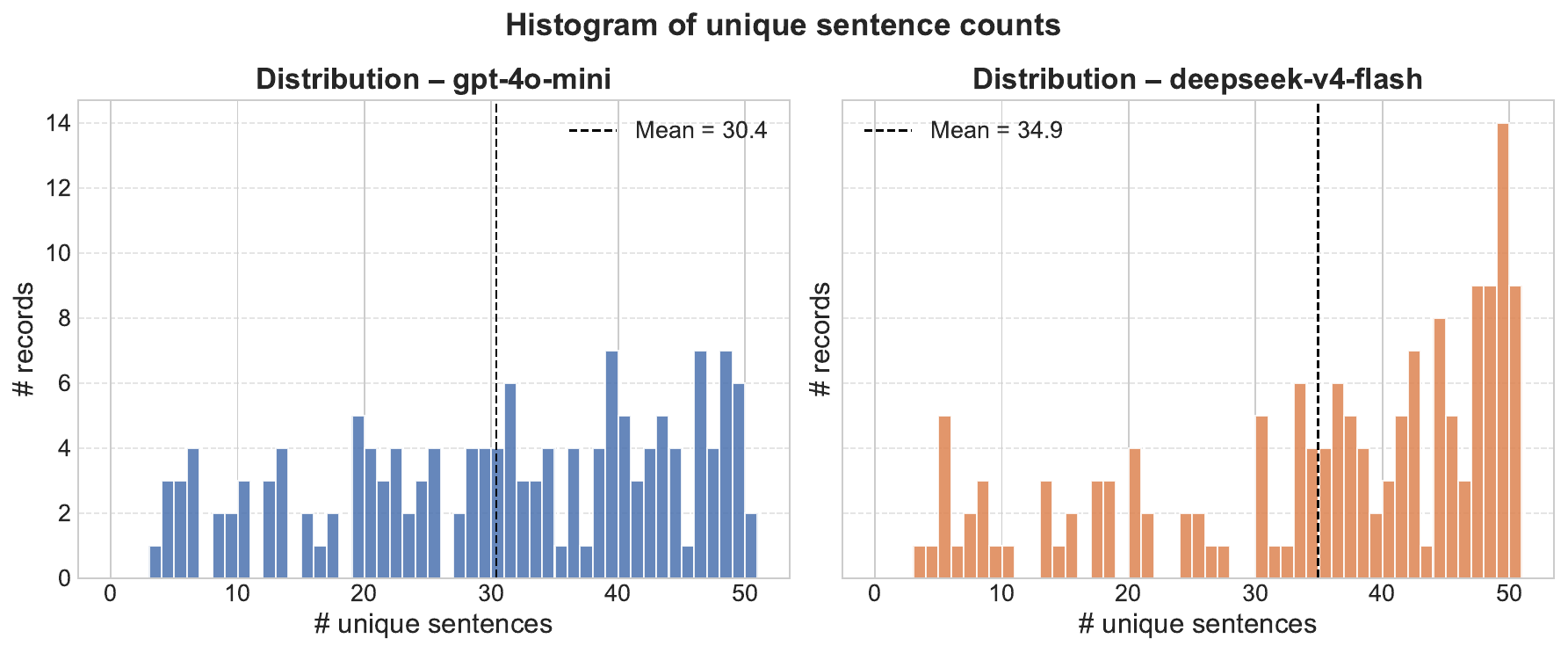}
    \vspace{-2em}
    \caption{Number of distinct sentences generated when alternating between GPT-4o mini and DeepSeek-V4 Flash on the BookSum dataset for 50 iterations.} 
    \label{llm_mix}
    \vspace{-1em}
\end{figure}

To better approximate heterogeneous real-world pipelines, we next introduce prompt heterogeneity across iterations by alternating prompts and LLMs.

Figure~\ref{prompt_mix} shows that prompt alternation increases the number of distinct outputs relative to a single fixed prompt under the same sampling-based configuration, but does not eliminate exact recurrences: some sentences still reappear across iterations. This setting remains \emph{Markovian} at the sentence level, but the kernel becomes \emph{time-inhomogeneous}: iteration $t$ uses $P^{(t)}_{M,\rho_t,d}$ rather than a single fixed $P_{M,\rho,d}$. Equivalently, the process can be cast as a time-homogeneous Markov chain on an augmented state space $(s,k)$ where $k$ indexes the active prompt (or prompt schedule). Round-trip translation can be interpreted as a structured instance of such heterogeneity, since each iteration composes distinct directional mappings (EN$\rightarrow \ell$ and $\ell\rightarrow$EN).

A similar pattern emerges when alternating between different LLMs. For instance, Figure~\ref{llm_mix} illustrates the case of alternating between GPT-4o mini and DeepSeek V4 Flash: the outputs of the LLMs are not unique, but some sentences still repeat.

\subsubsection{Paragraph-level iterative reprocessing}

Our primary experiments model the chain state at the sentence level. To examine whether analogous recurrence phenomena persist at larger granularities, we additionally run paragraph-level simulations on 450 paragraph seeds (150 per dataset), using \textit{GPT-4o-mini} with temperature $0.7$ and top-$p$ $0.9$ for $T=50$ transitions. Exact recurrence at the paragraph level (i.e., verbatim repetition of the full multi-sentence input) is uncommon within this horizon. However, recurrence remains pronounced at the sentence level: when we segment each paragraph output into sentences, individual sentence forms can reappear frequently across iterations, indicating that local attractor-like behavior can persist even when the state comprises multiple sentences. Some examples are provided in Table~\ref{long_text_example} in the Appendix.

To quantify exploration at this granularity, we compute a \emph{normalized diversity ratio}: the number of distinct sentence realizations observed over the iterative trajectory divided by the number of sentences in the original paragraph. Over 50 iterations, this ratio is 26.7 for BookSum, 19.7 for ScriptBase-alpha, and 24.2 for News2024, indicating that iterative paragraph-level reprocessing can generate substantial sentence-level variation even when full-paragraph exact recurrence is rare.

\subsubsection{Compared to Google Translator}

\begin{figure*}[t]
    \centering
    \includegraphics[width=\textwidth]{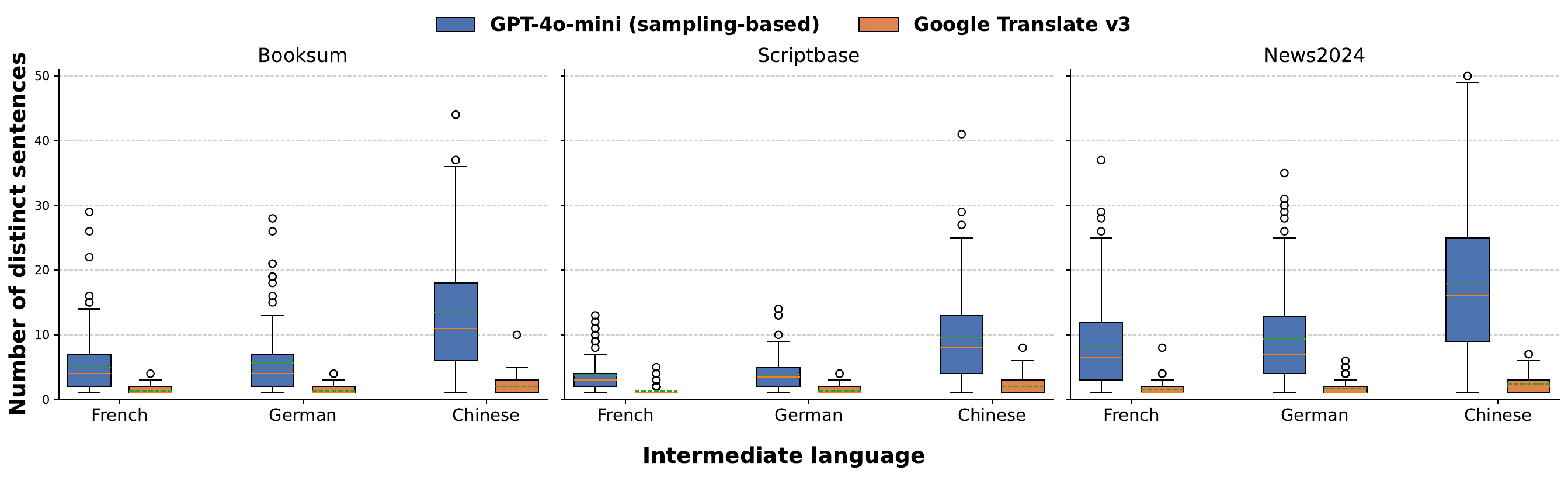}
    \vspace{-0.5em}
    \caption{Distinct-sentence counts under iterated round-trip translation for \textit{GPT-4o-mini} (sampling-based decoding) and \textit{Google Translate (v3)}.}
    \vspace{-1em}
    \label{gpt4o_gt}
\end{figure*}

We additionally instantiate iterative reprocessing through round-trip translation (EN$\rightarrow \ell \rightarrow$EN). Appendix Tables~\ref{translation_example} and~\ref{translation_example_2} provide qualitative examples of two representative finite-horizon behaviors: early entry into a fixed point and oscillation among a small set of closely related variants. 

We evaluate multiple bridge languages and compare sampling-based LLM translation with the production MT service \textit{Google Translate (v3)} as a baseline. Figure~\ref{gpt4o_gt} reports distinct-sentence counts under iterated round-trip translation for \textit{GPT-4o-mini} and \textit{Google Translate (v3)}. Unlike prompted LLM translation, which can exhibit substantial stochastic variation under sampling-based decoding, production MT services tend to behave nearly deterministically for fixed inputs. These results highlight that prompted LLM translation can induce substantially stronger surface-form variability under iterative reuse than a production MT service, even under a nominal meaning-preservation objective.

\subsubsection{Extended-horizon Behavior}

\begin{table}[t]
\centering
\small
\setlength{\tabcolsep}{4pt}
\begin{tabular}{lccc}
\toprule
 & ScriptBase & BookSum & News2024 \\
\midrule
\multicolumn{4}{l}{\emph{Distinct sentences}} \\
\quad $T{=}50$, mean & 24.1 & 24.2 & 30.7 \\
\quad $T{=}50$, median & 23.0 & 25.0 & 31.5 \\
\quad $T{=}500$, mean & 104.2 & 105.6 & 162.8 \\
\quad $T{=}500$, median & 90.0 & 71.5 & 125.0 \\
\midrule
\multicolumn{4}{l}{\emph{Final 50 iterations}} \\
\quad $\leq 3$ distinct & 7.3\% & 11.3\% & 5.3\% \\
\quad novel forms & 67.3\% & 66.0\% & 76.0\% \\
\bottomrule
\end{tabular}
\caption{Extended-horizon behavior under sampling-based decoding (GPT-4o-mini, $\tau=0.7$, top-p=0.9), 150 seed sentences per dataset. Distinct-sentence counts are cumulative over the first 50 and over all 500 iterations. The lower block reports the fraction of chains showing at most 3 distinct forms in the final 50 iterations, and the fraction still generating at least one form not seen earlier in the chain. Recurrence is exact string equality.}
\label{tab:horizon}
\end{table}

\begin{table}[t]
\centering
\small
\setlength{\tabcolsep}{4pt}
\begin{tabular}{lccc}
\toprule
 & ScriptBase & BookSum & News2024 \\
\midrule
\multicolumn{4}{l}{\emph{Estimated kernel $\hat{P}$}} \\
\quad chains & 143 & 141 & 134 \\
\quad core states & 11.0 & 10.1 & 10.3 \\
\quad $\hat{\pi}(\mathrm{tail})$ & 0.27 & 0.26 & 0.38 \\
\quad gap $1-|\lambda_2|$ & 0.32 & 0.30 & 0.41 \\
\midrule
\multicolumn{4}{l}{\emph{Held-out validation}} \\
\quad Kac $r_{\log}$ & 0.65 & 0.73 & 0.53 \\
\quad TV$(\hat{\pi})$ & 0.41 & 0.31 & 0.34 \\
\quad TV$_{\mathrm{base}}$ & 0.33 & 0.25 & 0.26 \\
\bottomrule
\end{tabular}
\caption{Transition kernels estimated on the core of each chain (states visited $\geq 5$ times), fitted on the first half of each chain and validated on the second. Chains whose fit-half core contains fewer than 3 states are excluded (7, 9 and 16 of 150). $\hat{\pi}(\mathrm{tail})$ is the stationary mass on the lumped tail state. Counts are smoothed by $\alpha=1/2$ before normalisation, which keeps $\widehat{P}$ strictly positive but inflates $1/\hat{\pi}_i$. Kac $r_{\log}$ is the log--log correlation between the prediction $1/\hat{\pi}_i$ and empirical return times in held-out data. The last two rows give the total-variation distance to held-out occupancy for $\hat{\pi}$ and for the naive baseline of reusing fit-half occupancy (median over chains).}
\label{tab:kernel}
\end{table}

To verify that our findings are not an artifact of the $T=50$ horizon, we conducted a set of extended-horizon checks across datasets. Table~\ref{tab:horizon} shows the  results with GPT-4o-mini under sampling-based decoding ($\tau=0.7$, top-p = 0.9) for $T=500$ iterations. Sentence-level diversity remains high; for example, more than two-thirds of the chains are still generating sentences that have never appeared before between steps 451 and 500.

We next estimate the transition kernel directly. Although the sentence state space is unbounded, the occupancy of each chain is concentrated on a small set of recurring sentences; we restrict each chain to this core, with every other sentence represented by a single tail state, and fit $\widehat{P}$ on the first half, evaluating on the second. Table~\ref{tab:kernel} reports the resulting estimates. Writing $\hat{\pi}$ for the stationary distribution of $\widehat{P}$, Kac's formula predicts a mean gap of $1/\hat{\pi}_i$ between successive visits to state $i$, and these predictions correlate with the observed gaps in the held-out half ($r_{\log}=0.53$--$0.73$). The stationary distribution itself, however, does not improve on reusing fit-half frequencies, and the spectral gap is narrow: over 500 iterations these chains have not settled into a stationary regime.

\section{Related Work}

\paragraph{Model collapse and iterative generation.}
A growing body of work investigates \emph{model collapse}, where iteratively training on synthetic data can degrade coverage of the underlying data distribution~\citep{shumailov2024ai,guo2023curious}. Follow-up studies analyze when collapse arises and propose mitigations such as mixing real and synthetic data or other data curation strategies~\citep{gerstgrasser2024model,seddik2024bad}, with broader concerns about downstream ``knowledge collapse'' in information ecosystems~\citep{peterson2025ai,wright2025epistemic}. Repeated LLM use can also form multi-step transformation chains (e.g., iterated paraphrasing or round-trip translation) that accumulate changes across calls, with observed distortion~\citep{perez2025llms,mohamed2025llm}. 

\paragraph{Sampling and diversity in LLM-generated content.} Sampling methods derived from the softmax distribution, including top-$k$~\citep{fan2018hierarchical}, top-$p$~\citep{holtzman2019curious}, and temperature-based sampling or logit suppression~\citep{chung2023increasing}, induce controlled randomness in generation, despite known limitations of softmax-based sampling in certain regimes~\citep{chang2022softmax}. Moreover, architectural and distributional constraints impose statistical regularities on outputs~\citep{mikhaylovskiy2025zipf}. A substantial body of work studies stylistic and diversity differences between LLM-generated and human-written text~\citep{munoz2024contrasting}. Several studies suggest that LLM-assisted generation can reduce downstream diversity, including in creative settings~\citep{padmakumar2023does,xu2025echoes,jiang2025artificial}. Most evaluations of LLM-generated content emphasize token-, phrase-, or corpus-level statistics~\citep{chung2023increasing,guo2024benchmarking,smith2025comprehensive}; however, sentence-level variation is a natural granularity for iterative rewriting and translation pipelines, providing a transparent and interpretable complement to existing metrics.

\paragraph{Language change and convergence in human--LLM interaction.}
Our work also relates to research on language change and cultural transmission~\citep{lieberman2007quantifying,kirby2007innateness,griffiths2007language,hamilton2016diachronic} and to studies of machine-mediated communication~\citep{brinkmann2023machine,arnon2024cultural}. In interaction settings, LLMs adapt to conversational partners~\citep{kandra2025llms,blevins2025language}. Users may learn and converge toward model language patterns over time~\citep{yakura2024empirical,geng2025impact}. We contribute an inference-time mechanism, iterative reprocessing under a fixed prompting setup, that provides a controlled lens on repeated reuse in these settings. 

\section{Distinction from Training-time Model Collapse}

\citet{shumailov2024ai} describe \emph{model collapse} as a training-time phenomenon in which repeated optimization on model-generated data can reduce coverage of the original data distribution. Subsequent analyses argue that the severity of this effect depends on strong assumptions and that it may be attenuated in more realistic training settings~\citep{schaeffer2025position}. Our setting is mechanistically distinct: we investigate \emph{inference-time} recursion under a fixed model, prompt, and decoding configuration, where an induced transformation operator is applied iteratively without any parameter updates.

Model collapse focuses on the parameters of the model, whereas we focus on the comparison between the input and output of the model. The behaviors we observe are attributable to properties of the induced transition kernel rather than to distributional contraction driven by learning. Moreover, nothing in the entropy behavior of these kernels rules out an increase in sentence-level diversity, in contrast to the degradation typically emphasized in training-time collapse accounts.

\section{Discussions and Conclusions}

LLMs are progressively integrated into text-processing pipelines, where their outputs can be fed into subsequent steps, either within a single workflow or reused across users. Many researchers are interested in exploring the differences between single-turn and multi-turn interactions in LLMs~\citep{li2025beyond,laban2025llms,huang2026llms}, whereas the scenarios considered in our paper can be regarded as independent sequential single-turn interactions. 

LLMs are frequently applied to tasks such as translation and rewriting. Although the behavior of LLMs in more settings requires further investigation, our experiments have already shown that repeated inference-time reprocessing is mechanistically distinct from training-time model-collapse phenomena. For example, Eq.~(\ref{eq:hx}) leaves open the possibility that iterative reprocessing increases sentence-level diversity when the input distribution is concentrated, since the induced kernels are not doubly stochastic.

The \emph{Markovian generation chains} we define naturally arise in the real world, for instance, through interactions among different users. Our Markov-chain framing provides a new perspective and a compact way to describe these phenomena and connect them to standard properties of stochastic operators (e.g., contraction of divergences under repeated application of a fixed kernel).

\section*{Limitations}
Real-world usage scenarios are far more complex than those in our simulations, and we also think that it is impossible to fully replicate every situation, for example, more LLMs, different prompts, and human involvement.

Our research has also been conducted primarily on English texts, and many types of English texts have not been covered. We chose the first sentence of the data to avoid the impression of selectively choosing sentences, but subsequent sentences may also exhibit different behavior.

Except when the temperature is set to 0, the same temperature values are not directly comparable across different models. We also do not provide an in-depth analysis of why different models show different performance, as this is beyond the scope of our work.

The kernels we estimate are restricted to each chain's core, with all remaining sentences represented by a single state, so they approximate rather than recover the induced kernel.

\section*{Acknowledgments}

This work benefited from funding from the French State, managed by the Agence Nationale de la Recherche, under the France 2030 program (grant reference ANR-23-IACL-0008). This research also received support from the ENS-PSL BeYs Chair in Data Science and Cybersecurity. We appreciate the thoughtful conversations and valuable suggestions from Gibbs Nwemadji.

\bibliography{emnlp}

\appendix

\newpage
\clearpage
\section{Use of LLMs}
LLMs were used to help improve the text, optimize the code, and assist with the analysis. Moreover, we took advantage of their search functionality as a complement to Google Search and also attempted to have LLMs assess potential issues in some of our arguments.

\section{Hardware.}
The experiments of Llama, Mistral and Qwen  were conducted on \textbf{8$\times$AMD MI210} GPUs, with a total compute allocation of 27 GPU-days.

\section{Token-level stochasticity and decoding}
\label{appendix_method_decoding}

Iterative trajectories depend on decoding-induced stochasticity~\citep{jang2016categorical}, in addition to system-level nondeterminism~\citep{he2025nondeterminism}. At generation step $t'$, let $z(\cdot \,;\, x,w_{<t'})$ denote the logits given input $x$ and prefix $w_{<t'}$. Temperature $\tau$ induces

\begin{equation}
\begin{split}
& \pi_{\tau}(w_{t'} \mid x, w_{<t'}) \\
&\quad = \frac{\exp(z(w_{t'}; x,w_{<t'})/\tau)}{\sum_{v}\exp(z(v; x,w_{<t'})/\tau)} \,,
\end{split}
\end{equation}
where larger $\tau$ typically increases randomness~\citep{holtzman2019curious}. Top-$k$, nucleus sampling (top-$p$), and logit controls further truncate or reweighs this distribution (with limitations in some regimes~\citep{chang2022softmax}). The probability of a length-$n$ sequence is

\begin{equation}
\Pi_{\tau}(w_{1:n}\mid x)=\prod_{{t'}=1}^{n}\pi_{\tau}(w_{t'}\mid x,w_{<t'}) \,.
\label{eq_sentence}
\end{equation}
These token-level choices induce the sentence-level transition kernel in Eq.~(\ref{eq:mgc_step}) and can compound under iteration. Even when alternative decoding objectives are proposed, such as multi-token prediction~\citep{gloeckle2024better}, the core stochastic selection mechanism remains central to typical deployments.

\section{Contraction of relative entropy}
\label{appendix_KL}

KL divergence is defined as,
\begin{equation}
D_{KL}(X \| Y) = \sum_i X_i \log \left( \frac{X_i}{Y_i} \right) \,.
\end{equation}

Similarly,
\begin{equation}
\begin{split}
& D_{\mathrm{KL}}(X\mathbf{P} \| Y\mathbf{P}) \\
&\quad = \sum_j (X\mathbf{P})_j
\log \left( \frac{(X\mathbf{P})_j}{(Y\mathbf{P})_j} \right),
\end{split}
\end{equation}
where $(X\mathbf{P})_j = \sum_i X_i P_{ij}$ and $(Y\mathbf{P})_j = \sum_i Y_i P_{ij}$.
Assume $(Y\mathbf{P})_j>0$ whenever $(X\mathbf{P})_j>0$ (otherwise the divergence is
infinite and the inequality holds trivially). By the log-sum inequality, for each $j$,
\begin{equation}
\begin{split}
\biggl(\sum_i X_i P_{ij}\biggr)
&\log\frac{\sum_i X_i P_{ij}}{\sum_i Y_i P_{ij}} \\
&\le \sum_i X_i P_{ij}\log\frac{X_i P_{ij}}{Y_i P_{ij}},
\end{split}
\end{equation}
with the convention that terms with $X_iP_{ij}=0$ contribute $0$. Summing over $j$ gives
\begin{equation}
\begin{split}
& D_{\mathrm{KL}}(X\mathbf{P} \| Y\mathbf{P}) \\
&\quad\le \sum_j \sum_i X_i P_{ij}\log\frac{X_i P_{ij}}{Y_i P_{ij}} \\
&\quad= \sum_i X_i \log\frac{X_i}{Y_i}\sum_j P_{ij}.
\end{split}
\end{equation}
Since $\mathbf{P}$ is row-stochastic, $\sum_j P_{ij}=1$, hence
\begin{equation}
D_{\mathrm{KL}}(X\mathbf{P} \| Y\mathbf{P}) \le D_{\mathrm{KL}}(X \| Y).
\end{equation}

\section{Prompts}
\label{appendix_prompts}

This section lists the prompt templates used in the main experiments and ablations. In all listings, \texttt{\{content\}} is replaced with the current input text at iteration $t$, and \texttt{\{target\_lang\}} denotes the chosen bridge language for round-trip translation. Unless stated otherwise, the same template is reused across iterations.

\paragraph{Tasks.}
\textbf{Iterative rephrasing.} Using rephrasing prompt $\rho_{\mathrm{reph}}$, we iterate
$s^{(t+1)} \sim \mathcal{T}_{M,\rho_{\mathrm{reph}},d}(\cdot \mid s^{(t)})$.
\textbf{Iterated round-trip translation.} For bridge language $\ell$ with prompts $\rho_{\mathrm{tr}}$, one iteration applies $\mathcal{T}^{\text{EN}\to \ell \to \text{EN}}_{M,\rho_{\mathrm{tr}},d}$ (Eq.~\ref{eq:rt_operator}):
\[
s^{(t)}_{\text{EN}} \xrightarrow{\text{EN}\to \ell} u^{(t)}_{\ell} \xrightarrow{\ell\to \text{EN}} s^{(t+1)}_{\text{EN}}.
\]

\paragraph{Rephrasing prompts.}
Listing~\ref{prompt_example} is the main meaning-preserving rephrasing template, and Listing~\ref{prompt_ablation} is a shorter ablation variant used in the prompt-sensitivity study.

\begin{lstlisting}[caption={Prompt for rephrasing},label=prompt_example]
    "Given a passage, rephrase it while preserving all the original meaning and without losing any context.\n"
    "Do not write an introduction or a summary. Return only the rephrased passage.\n\n"
    "Rephrase the following text:\n{content}"
\end{lstlisting}

\begin{lstlisting}[caption={Prompt for rephrasing (ablation)},label=prompt_ablation]
    "Rephrase the following text:\n{content}"
\end{lstlisting}

\paragraph{Translation prompts.}
For round-trip translation, we use direction-specific templates (EN$\to\ell$ and
$\ell\to$EN). The current input text is appended after the instruction, and
\texttt{\{target\_lang\}} specifies the intermediate language $\ell$.

\begin{lstlisting}[caption={Prompt for translation}]
    "Translate the following English text into {target_lang}:"
    "Translate the following {target_lang} text into English:"
\end{lstlisting}

\section{Additional Results}
\label{app:appendix_addition_results}

This section provides supplementary qualitative trajectories and full metric plots referenced in the main text. We include (i) iterative rephrasing examples under sampling, (ii) round-trip translation examples, (iii) similarity-metric trajectories for additional datasets/decoding regimes, and (iv) full length--diversity correlation outputs.

\paragraph{Iterative rephrasing under sampling.}
Table~\ref{rephrasing_example_temperature} shows representative trajectories
under sampling-based decoding ($\tau=0.7$, top-$p=0.9$), illustrating how some
chains exhibit continued drift while others still enter short cycles.

\paragraph{Iterated round-trip translation examples.}
Tables~\ref{translation_example} and~\ref{translation_example_2} provide illustrative round-trip translation chains (EN$\rightarrow$French$\rightarrow$EN), highlighting early stabilization and occasional alternation between near-identical variants.

\begin{figure*}[t]
    \centering
    \includegraphics[width=\textwidth]{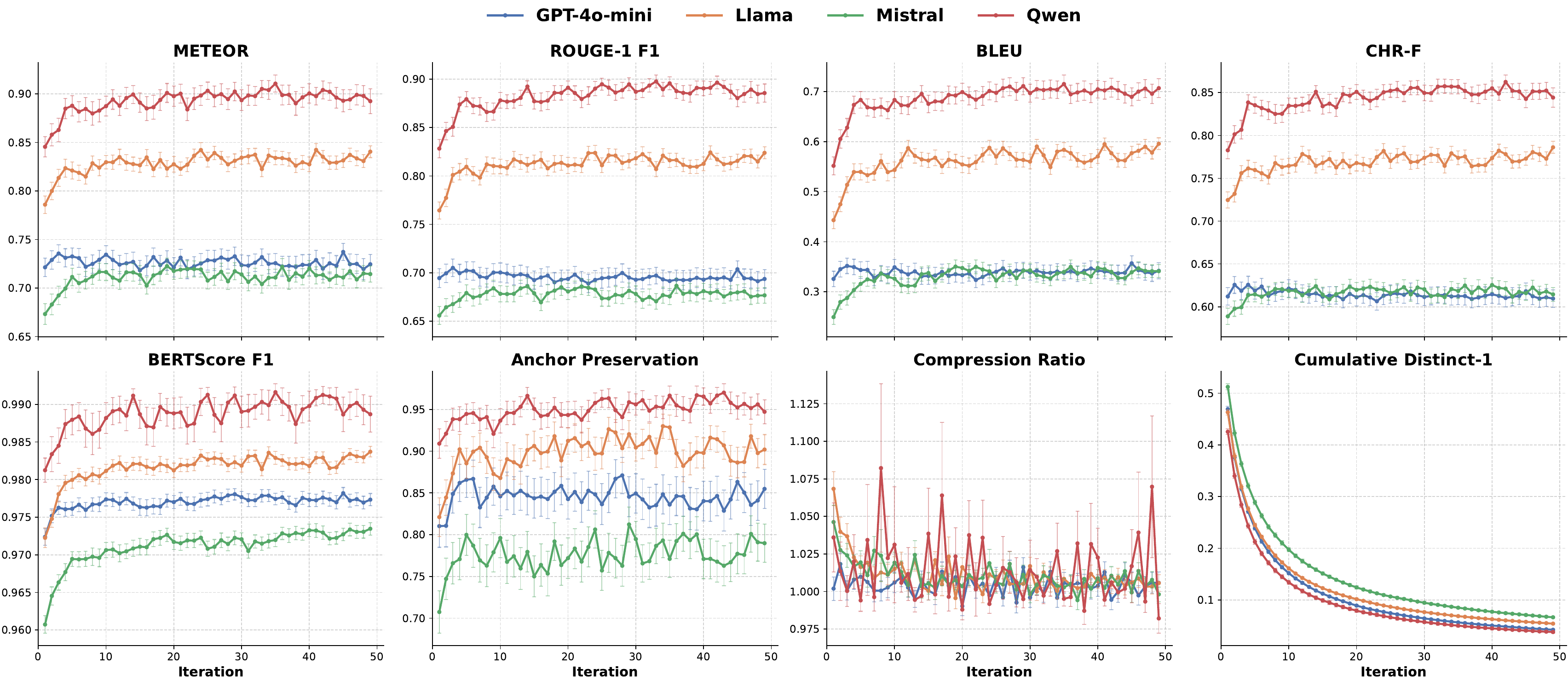}
    \caption{Evolution of text similarity metrics across 50 rephrasing iterations for the News2024 dataset using sampling-based decoding.}

    \label{fig:news2024_models_comparison_subplot_sampling}
\end{figure*}

\begin{figure*}[t]
    \centering
    \includegraphics[width=\textwidth]{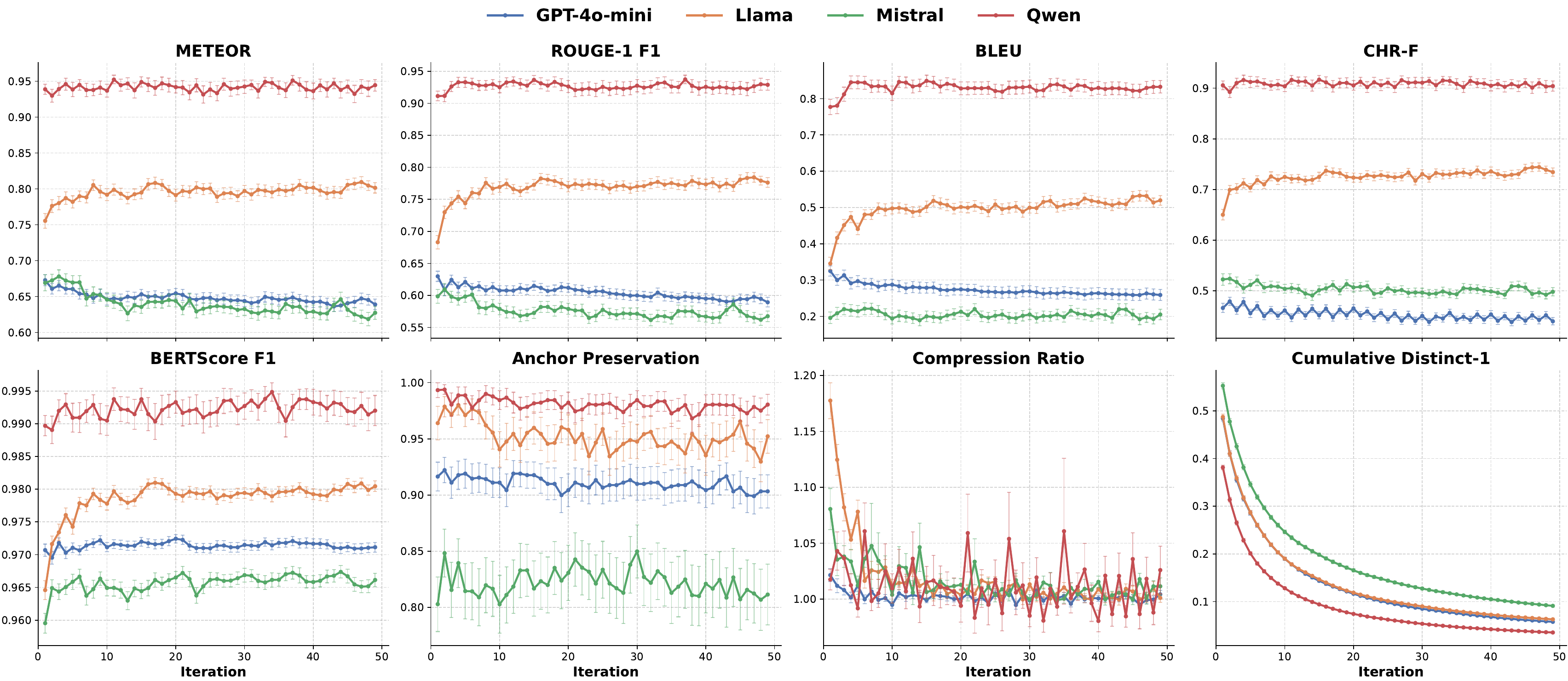}
    \caption{Evolution of text similarity metrics across 50 rephrasing iterations for the ScriptBase dataset using sampling-based decoding.}

    \label{fig:scriptbase_models_comparison_subplot_sampling}
\end{figure*}

\begin{figure*}[t]
    \centering
    \includegraphics[width=\textwidth]{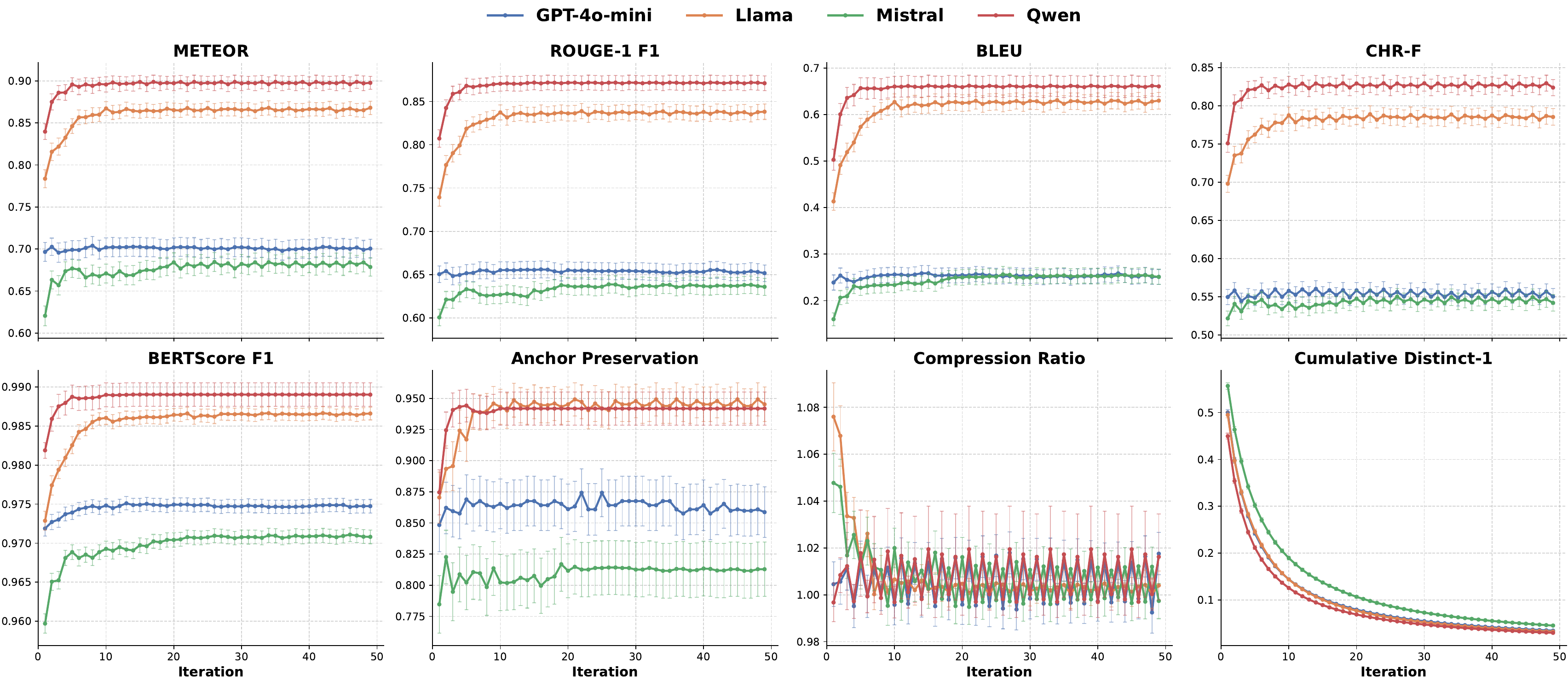}
    \vspace{-0.5em}
    \caption{Evolution of text similarity metrics across 50 rephrasing iterations for the BookSum dataset using greedy decoding.}
    \label{fig:booksum_models_comparison_subplot_greedy}
    \vspace{-1.5em}
\end{figure*}

\begin{figure*}[t]
    \centering
    \includegraphics[width=\textwidth]{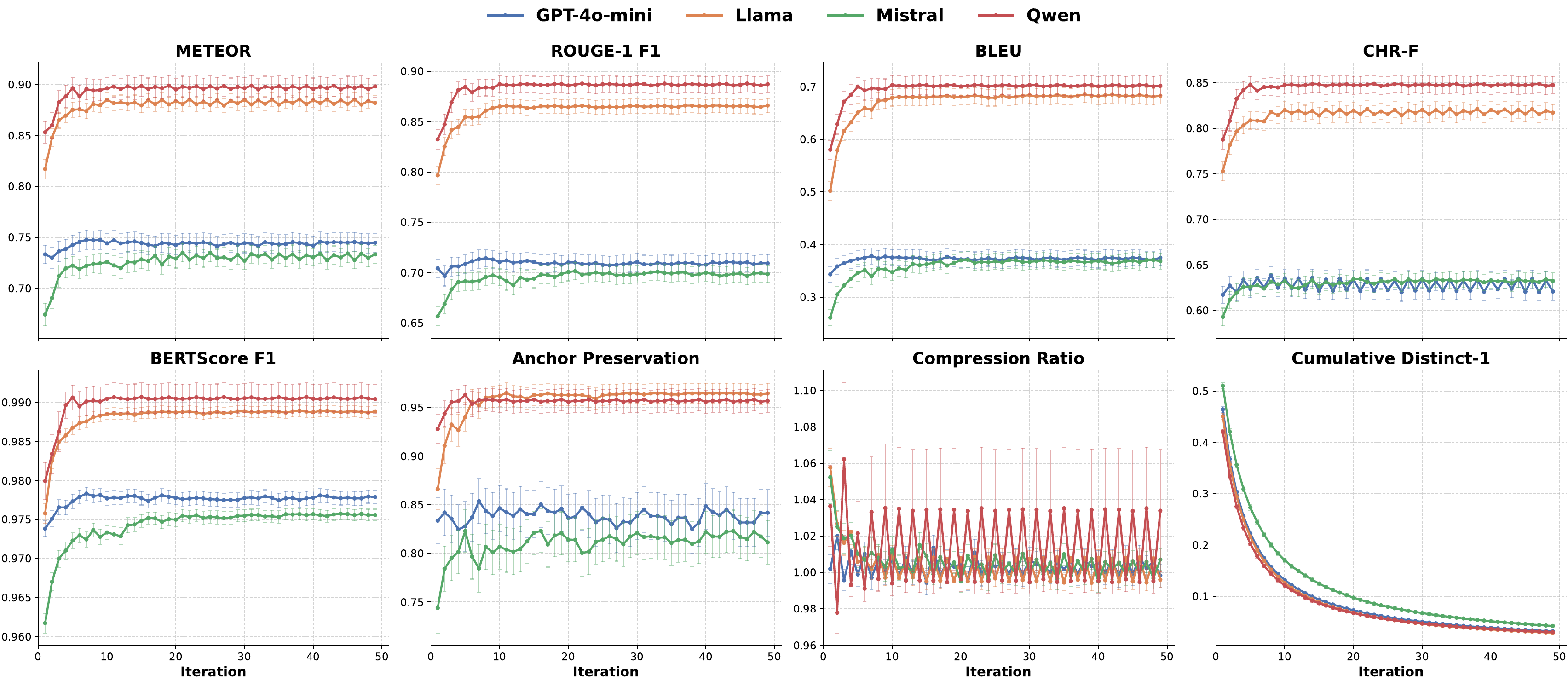}
    \caption{Evolution of text similarity metrics across 50 rephrasing iterations for the News2024 dataset using greedy decoding.}

    \label{fig:news2024_models_comparison_subplot_greedy}
\end{figure*}

\begin{figure*}[t]
    \centering
    \includegraphics[width=\textwidth]{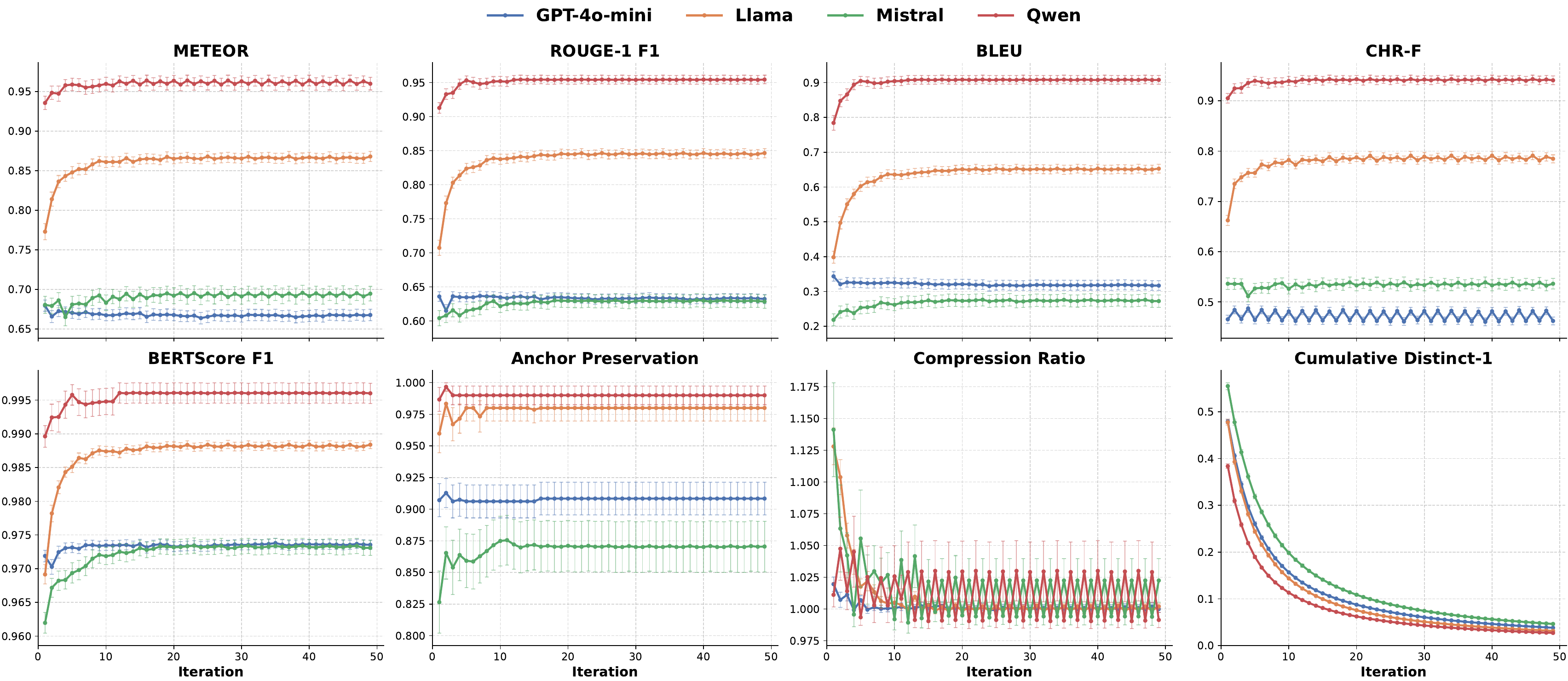}
    \caption{Evolution of text similarity metrics across 50 rephrasing iterations for the ScriptBase dataset using greedy decoding.}

    \label{fig:scriptbase_models_comparison_subplot_greedy}
\end{figure*}

\paragraph{Similarity dynamics across iterations.}
Figures~\ref{fig:news2024_models_comparison_subplot_sampling} and \ref{fig:scriptbase_models_comparison_subplot_sampling} report the analogous plots under sampling-based decoding. Figures~\ref{fig:booksum_models_comparison_subplot_greedy}--\ref{fig:scriptbase_models_comparison_subplot_greedy} extend the greedy-decoding analysis to additional datasets. As in the main text, each point compares iteration $t$ to iteration $t-1$ using the eight metrics.

\paragraph{Full length--diversity correlation results.}
Table~\ref{tab_correlation} reports Pearson correlations between seed length and diversity, along with $p$-values, linear-fit $R^2$, and slopes for each model/decoding setting and dataset.

\newpage

\begin{table*}[t]
  \centering
  \begin{tabularx}{\linewidth}{>{\raggedright\arraybackslash}p{1cm} X X}
    \toprule
    Round          & \textbf{Llama-3.1-8B-Instruct} & \textbf{Qwen2.5-7B-Instruct} \\
    \midrule
     0     &    We begin with a prologue.  &  We begin with a prologue.  \\
     1  &  The narrative commences with a prologue.   &  We start with a prologue.  \\
     2  & The story begins with a prologue that sets the scene. &  We begin with a prologue. \\
     3  & The narrative commences with a prologue that establishes the backdrop for the story. &  We start with a prologue. \\
     5  & The story starts with a prologue, which establishes the setting and lays the groundwork for the events that unfold in the narrative. &  We start with a prologue. \\
     10  & The narrative begins with a prologue, which sets the scene and lays the groundwork for the events that unfold in the story that follows. & We begin with a prologue. \\
     20  & The narrative begins with a prologue, establishing the groundwork and setting the scene for the events that will unfold throughout the story. & We begin with a prologue. \\
     30  & The narrative begins with a prologue that sets the scene and lays the groundwork, providing a solid foundation for the events that follow to unfold and develop. & We begin with a prologue. \\
     40  & The narrative begins with a prologue that sets the scene, provides essential background information, and lays the groundwork, establishing a solid foundation that allows the subsequent events to unfold in a logical and cohesive manner. & We begin with a prologue. \\
     49 &  The story starts with a prologue that lays the groundwork by providing key background information and establishing the setting, thus creating a solid foundation that allows the subsequent events to unfold in a logical and coherent manner. & We start with a prologue. \\
     50 & The story begins with a prologue that sets the stage by offering essential background information and establishing the setting, thereby laying a solid foundation that enables subsequent events to unfold in a logical and coherent way. & We begin with a prologue. \\
    \bottomrule
  \end{tabularx}
\caption{Examples of repeated rephrasing by different LLMs (temperature=0.7, top-p=0.9).}
\label{rephrasing_example_temperature}
\end{table*}

\begin{table*}[t]
\centering
\begin{tabularx}{\textwidth}{>{\raggedright\arraybackslash}p{1cm} X X}
\toprule
Round & French  &  English \\
\midrule
0 &  &  We begin with a prologue. \\
1 & Nous commençons par un prologue. & We begin with a prologue.  \\
2 & Nous commençons par un prologue. & We begin with a prologue. \\
3 & Nous commençons par un prologue. & We begin with a prologue. \\
4 & Nous commençons par un prologue. & We begin with a prologue. \\
\bottomrule
\end{tabularx}
\caption{Examples of iterative translation via GPT-4o-mini (temperature=0.7, top-p=0.9).}
\label{translation_example}
\end{table*}

\begin{table*}[t]
\centering
\begin{tabularx}{\linewidth}{>{\raggedright\arraybackslash}p{1cm} X X}
\toprule
Round & French  &  English \\
\midrule
0 &  &  Elizabeth \textbf{reads} through Darcy's letter with a mixture of emotions. \\
1 & Elizabeth \textbf{lit} la lettre de Darcy avec un mélange d'émotions. & Elizabeth read Darcy's letter with a mix of emotions.  \\
2  & \textit{Elizabeth a lu la lettre de Darcy avec un mélange d'émotions.}  &  \textit{Elizabeth read Darcy's letter with a mix of emotions.} \\
\vdots  & \vdots & \vdots \\
11  & Elizabeth \textbf{lut} la lettre de Darcy avec un mélange d'émotions. & Elizabeth read Darcy's letter with a \textbf{mixture} of emotions. \\
12  & Elizabeth a lu la lettre de Darcy avec un mélange d'émotions. & Elizabeth read Darcy's letter with a \textbf{mixture} of emotions. \\
\vdots  & \vdots & \vdots \\
 19 & Elizabeth \textbf{lut} la lettre de Darcy avec un mélange d'émotions. &  Elizabeth read Darcy's letter with a mix of emotions.\\
\bottomrule
\end{tabularx}
\caption{Examples of iterative translation via GPT-4o-mini (temperature=0.7, top-p=0.9). The omitted lines are all identical to the content in the second round (the italicized sentences). The similar scenario also occurs after the 19th round.}
\label{translation_example_2}
\end{table*}

\begin{table*}[t]
\centering
\begin{tabularx}{\linewidth}{@{}>{\raggedright\arraybackslash}X >{\centering\arraybackslash}p{2cm}@{}}
\toprule
\textbf{Sentence} & \textbf{Frequency} \\
\midrule
Hooray for best friends! & 50 \\
I don't think so. & 50 \\
 You're the ones being banished!" & 50 \\
Sue now has two children and is expecting a third. & 46 \\
Suddenly, the ghost appears. & 41 \\
Jo continues to write short romantic stories for the newspaper while secretly working on a novel. & 34 \\
Just as she is about to leave, the Westons arrive in their carriage. & 33 \\
Cunegonde resists the advances of both men. & 32 \\
 They have several children, with Charlotte being the eldest; she is practical, sensible, and intelligent. & 32 \\
Madame Defarge passes by and acknowledges them with a nod. & 31 \\
\bottomrule
\end{tabularx}
\caption{Highest recurrence frequency observed sentences within the 50-iteration paragraph-level runs (\textit{GPT-4o-mini}, sampling-based decoding, prompt P1) on 150 BookSum paragraphs.}
\label{long_text_example}
\end{table*}


\clearpage

\clearpage

\begin{table*}[t]
\centering
\begin{tabular}{l l c c c c}
\toprule
Dataset & Model & r & p & R² & Slope \\
\midrule
\multirow{8}{*}{booksum} 
& Llama greedy & 0.171 & 3.611e-02 & 0.029 & 0.107 \\
& Llama sample & 0.166 & 4.272e-02 & 0.027 & 0.191 \\
& Mistral greedy & 0.324 & 5.135e-05 & 0.105 & 0.350 \\
& Mistral sample & 0.438 & 2.046e-08 & 0.192 & 0.594 \\
& Qwen greedy & 0.353 & 9.427e-06 & 0.125 & 0.125 \\
& Qwen sample & 0.494 & 1.330e-10 & 0.244 & 0.616 \\
& GPT greedy & 0.476 & 7.658e-10 & 0.226 & 0.261 \\
& GPT sample & 0.638 & 1.542e-18 & 0.407 & 1.054 \\
\hline
\multirow{8}{*}{scriptbase} 
& Llama greedy & -0.174 & 3.338e-02 & 0.030 & -0.238 \\
& Llama sample & 0.030 & 7.150e-01 & 0.001 & 0.048 \\
& Mistral greedy & 0.073 & 3.731e-01 & 0.005 & 0.124 \\
& Mistral sample & -0.021 & 8.012e-01 & 0.000 & -0.050 \\
& Qwen greedy & 0.094 & 2.510e-01 & 0.009 & 0.056 \\
& Qwen sample & 0.157 & 5.479e-02 & 0.025 & 0.272 \\
& GPT greedy & 0.105 & 2.017e-01 & 0.011 & 0.109 \\
& GPT sample & 0.231 & 4.471e-03 & 0.053 & 0.727 \\
\hline
\multirow{8}{*}{news2024} 
& Llama greedy & 0.187 & 2.216e-02 & 0.035 & 0.107 \\
& Llama sample & 0.114 & 1.659e-01 & 0.013 & 0.081 \\
& Mistral greedy & 0.111 & 1.757e-01 & 0.012 & 0.118 \\
& Mistral sample & 0.197 & 1.585e-02 & 0.039 & 0.141 \\
& Qwen greedy & 0.199 & 1.454e-02 & 0.040 & 0.058 \\
& Qwen sample & 0.373 & 2.600e-06 & 0.139 & 0.477 \\
& GPT greedy & 0.367 & 3.765e-06 & 0.135 & 0.203 \\
& GPT sample & 0.385 & 1.116e-06 & 0.149 & 0.525 \\
\bottomrule
\end{tabular}
\caption{Results of the Correlation Analysis. $r$ represents the Pearson correlation coefficient, and $R^2$ represents the coefficient of determination.}
\label{tab_correlation}
\end{table*}

\end{document}